\documentclass[conference]{IEEEtran}
\IEEEoverridecommandlockouts
\usepackage{cite}
\usepackage{amsmath,amssymb,amsfonts}
\usepackage{algorithm}
\usepackage{algorithmic}

\usepackage{graphicx}
\usepackage{subfigure}

\usepackage{textcomp}
\usepackage{xcolor}
\usepackage{ulem}
\usepackage{multirow}
\usepackage{booktabs}
\usepackage{pifont}
\usepackage{balance}

\def\BibTeX{{\rm B\kern-.05em{\sc i\kern-.025em b}\kern-.08em
    T\kern-.1667em\lower.7ex\hbox{E}\kern-.125emX}}
    
\begin{document}
\title{\LARGE{Privacy Inference-Empowered Stealthy Backdoor Attack \\ on Federated Learning under Non-IID Scenarios}

\author{
    \IEEEauthorblockN{Haochen Mei\IEEEauthorrefmark{1}, Gaolei Li \IEEEauthorrefmark{1}\IEEEauthorrefmark{2}, Jun Wu\IEEEauthorrefmark{3}, Longfei Zheng\IEEEauthorrefmark{4}}
    \IEEEauthorblockA{\IEEEauthorrefmark{1} School of Electronic Information and Electrical Engineering, Shanghai Jiao Tong University, Shanghai, China}
    \IEEEauthorblockA{\IEEEauthorrefmark{2} Shanghai Key Laboratory of Integrated Administration Technologies for Information Security, Shanghai, China}
    \IEEEauthorblockA{\IEEEauthorrefmark{3} Graduate School of Information, Production and Systems, Waseda University, Fukuoka, Japan.}
    \IEEEauthorblockA{\IEEEauthorrefmark{4} Ant Group, China}
    \IEEEauthorblockA{\{sky-silhouette, gaolei\_li\}@sjtu.edu.cn, junwu@aoni.waseda.jp, zlf206411@antgroup.com}
}

}

\maketitle
\pagestyle{empty}

\begin{abstract}
Federated learning (FL) naturally faces the problem of data heterogeneity in real-world scenarios, but this is often overlooked by studies on FL security and privacy. On the one hand, the effectiveness of backdoor attacks on FL may drop significantly under non-IID scenarios. On the other hand, malicious clients may steal private data through privacy inference attacks. Therefore, it is necessary to have a comprehensive perspective of data heterogeneity, backdoor, and privacy inference. 
In this paper, we propose a novel privacy inference-empowered stealthy backdoor attack (PI-SBA) scheme for FL under non-IID scenarios. 
Firstly, a diverse data reconstruction mechanism
based on generative adversarial networks (GANs) is proposed
to produce a supplementary dataset, which can improve the
attacker’s local data distribution and support more sophisticated
strategies for backdoor attacks.
Based on this, we design a source-specified backdoor learning (SSBL) strategy as a demonstration, allowing the adversary to arbitrarily specify which classes are susceptible to the backdoor trigger. Since the PI-SBA has an independent poisoned data synthesis process, it can be integrated into existing backdoor attacks to improve their effectiveness and stealthiness in non-IID scenarios. Extensive experiments based on MNIST, CIFAR10 and Youtube Aligned Face datasets demonstrate that the proposed PI-SBA scheme is effective in non-IID FL and stealthy against state-of-the-art defense methods.

\end{abstract}


\section{Introduction}
\label{intro}
Federated learning (FL) provides a new paradigm for cooperative learning, which allows multiple parties to jointly train a global model without sharing local private data \cite{FL}. 
While providing better privacy protection, FL also exposes a larger attack surface than centralized learning. One of the most typical threats is from malicious clients who can launch poisoning attacks on the global model by uploading well-crafted local updates. The backdoor attack as a targeted poisoning attack is notoriously stealthy and threatening, which aims to induce the model to misbehave under specific circumstances while maintaining the original performance on the main task \cite{survey1}.

\begin{figure}[t]
\centering
\includegraphics[scale=0.42]{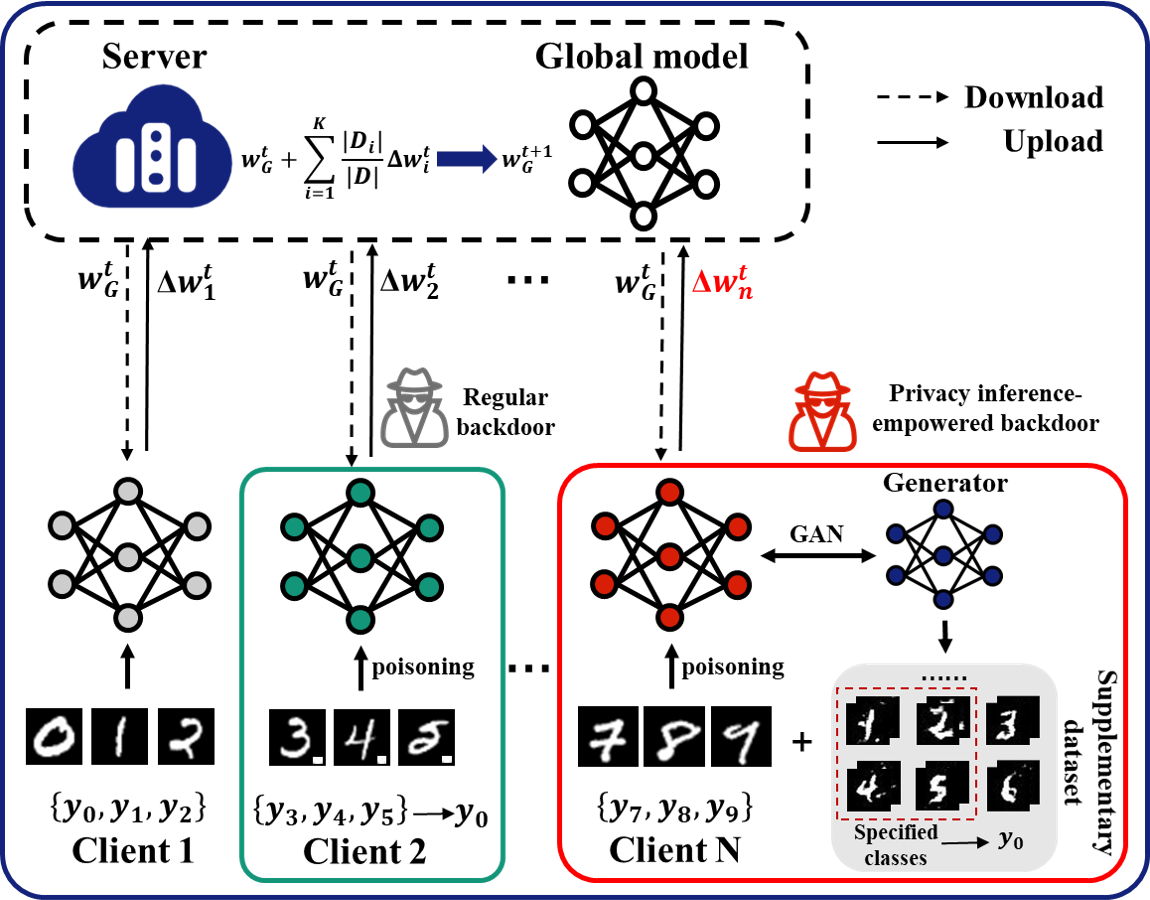}
\caption{An overview of FL with client-side backdoor attackers under the label distribution non-IID scenario.}
\label{fig1}

\end{figure}
Since proposed in \cite{badnets}, backdoor attacks have been extensively studied in centralized deep neural networks (DNNs) and many powerful variants have emerged \cite{yan2021deep,yang2022propagable}.
For FL, the migration of backdoor attacks seems to be a simple and straightforward task. For example, initiated by the attempt of Bagdasaryan et al. \cite{ba1}, a series of studies demonstrate the possibility of backdooring FL in various ways \cite{ba2,tailattack,poisongan,multitentacle}.
However, maintaining the backdoor effectiveness can be quite a challenge in such a complex distributed system. Here we list two main factors:
\begin{itemize}
    \item First, the updates from benign clients can weaken the impact of poisonous updates (especially when the proportion of malicious clients is quite small), and it makes the global model quickly forget the backdoor \cite{toxin}. To negate the benign updates' effect, an explicit amplification strategy has been widely used \cite{ba1,ba2,poisongan}. We also draw on this approach in the proposed scheme and further evaluate its effectiveness in section \ref{experiment}. 
    
    
    \item Second, a more serious but often overlooked issue is that data heterogeneity can greatly reduce the effectiveness of backdoor attacks in FL. Data heterogeneity can lead to the overfitting of local models on the skewed local data so that the backdoor features will be suppressed. What's worse, the difference between local malicious data distribution and global data distribution has a significant impact on the backdoor effectiveness \cite{curse}, which must be considered when designing the attack strategy. 
\end{itemize}

Through further analysis, we found that the existing backdoor attacks on FL have the following limitations: \textbf{\ding{172}} The issue of non-iid data mentioned above is not considered. Although we note that \cite{poisongan} attempted to alleviate the attacker's data absence by introducing the GAN-based inference method as \cite{deepmodel}, it can only restore the data of a single class, and only work in the special case where all class members are similar.
\textbf{\ding{173}} The triggers pattern used to control the backdoor behavior is missing in most existing so-called backdoor attacks in FL, which, however, is usually an important component in backdoor attacks on centralized DNNs \cite{survey3}. We also note that a novel semantic backdoor is proposed in \cite{ba1} (e.g., using a striped wall background as a semantic trigger for ``car" images), but this will greatly limit the attacker's flexibility. We speculate that this is due to the fact that the features of explicit trigger patterns are prone to failure in the non-IID scenario.

Therefore, to help break through the limitations faced by backdoor attackers in the non-IID FL, we propose a privacy inference-empowered stealthy backdoor attack (PI-SBA) scheme leveraging a GANs-based diverse data inference mechanism \cite{GAN}, which produces a supplementary dataset consisting of diverse generated samples to reduce the gap between local and global data distributions.
To make it clear, in this work, we mainly focus on a typical non-IID scenario---the label distribution skew \cite{noniid1,noniid2} in cross-silo FL \cite{survey1}, where a handful of clients (usually 10 to 100) cooperate to train a shared model with much larger datasets. As is shown in Figure \ref{fig1}, each client (including the attacker) only has access to a subset of all data classes. It is a common case in FL systems composed of large organizations such as hospitals or banks that own data with different labels from different sources.

Based on this, we further design a more flexible and insidious source-specified backdoor learning (SSBL) strategy that can arbitrarily specify which classes are susceptible and which are not.
It should be noted that the SSBL strategy not only demonstrates that our GANs-based diverse data inference mechanism can assist attackers to carry out more sophisticated backdoor attacks despite the absence of relevant original data, but also corresponds to a realistic threat model. 
For example, in an FL system consisting of a group of commercial companies with different user groups, for the purpose of vicious competition, the attacker only compromises the user groups of the rival companies, leaving the rest immune, which is more insidious than injecting a generic backdoor. 

We believe that our GANs-based diverse data inference mechanism as well as the SSBL strategy is not an isolated work, but can be integrated into existing backdoor attacks (.e.g, \cite{ba1,dba,canyou}) on FL to help improve their effectiveness and concealment under non-IID scenarios. The feasibility and effectiveness of the proposed PI-SBA scheme validate that \textbf{Inadvertently-leaked private data can be used to backdoor the FL models}. This may lead to benign users in FL systems being stigmatized as backdoor attackers.

Our main contributions in this work can be summarized as follows:
\begin{itemize}
    \item We propose a novel privacy inference-empowered backdoor attack (PI-SBA) scheme for non-IID FL, leveraging a GANs-based diverse data inference mechanism to improve the local data distribution for malicious clients and enhance the attack effectiveness in non-IID FL.
 
    \item A source-specified backdoor learning (SSBL) strategy is designed to allow malicious clients to attack any specified classes despite the absence of relevant data, which presents higher feasibility and flexibility, and is compatible with existing backdoor attacks to improve stealthiness.
   
    \item We conduct extensive experiments on MNIST, CIFAR10, and Youtube Aligned Face (YAF) datasets \cite{face} to comprehensively evaluate the performance of the proposed scheme. It is demonstrated that our methods can increase the attack success rate by 20\%--60\% for regular backdoor attacks in non-IID FL and can successfully evade two state-of-the-art defense methods.
\end{itemize}

\section{Related Work} \label{sec2}
\subsection{Backdoor Attacks in FL}
Bagdasaryan et al. \cite{ba1} proposed the first backdoor attack against FL, which selects specific semantics in the original data as the triggers for backdoor training and replaces the global model with the backdoored one by manipulating the model updates. 
Wang et al. \cite{tailattack} proposed an edge-case backdoor attack and first theoretically verified that if adversarial example attacks are effective against models in FL, so are backdoor attacks. Zhang et al. \cite{poisongan} introduced a generative poisoning attack called PoisonGAN for situations where the attacker does not have access to the training data, and reproduced the backdoor attack in \cite{ba1}. Considering multiple colluding malicious clients, Xie et al. \cite{dba} proposed to utilize a composite global trigger formed by several local triggers to conduct a distributed backdoor attack (DBA) on FL. Similarly, Gong et al. \cite{ba8} proposed to use model-agnostic triggers to increase the attack success rate of DBA. Besides, to improve the backdoor persistence in FL, Zhang et al. \cite{toxin} proposed a simple but efficient method called Neurotoxin to slightly modify model parameters during the FL training process. 



\subsection{Privacy Inference Attacks in FL}
To alleviate the problem of missing data for backdoor attackers in non-IID scenarios, we also investigate privacy inference attacks in FL. 
Melis et al. \cite{exploiting} demonstrated that the model updates could leak unintended information about clients' training data, and such leakage risk makes it possible for server-side attackers to reconstruct the original training data from collected updates.
Wang et al.\cite{userlevel} devised a framework incorporating the generative adversarial network (GAN) with a multitask discriminator to simultaneously discriminates category, reality, and client identity of input samples. However, server-side privacy inference attacks require higher attacker capabilities. 
Hitaj et al. \cite{deepmodel} extended the model inversion attack as a client-side privacy inference attack using the GAN to generate the targeted class representatives in a collaborative learning system. This attack exploited the real-time nature of the FL system and regards the global model as a discriminator to train the generator. However, considering the attack assumption and effect comprehensively, the above methods can not be directly applied to launch an effective backdoor attack on FL under non-IID data scenarios. 

\section{Proposed Method} 
\label{sec3}
\subsection{Preliminaries}
\subsubsection{Cross-silo FL with Label Distribution Non-IID Data} \label{pre1}
Cross-silo FL refers to the collaborative model training involving several (usually within one hundred \cite{survey1}) large organizations or companies. Here we consider a $N$ class classification problem on the dataset $D=\{(x,y)\}$, where the data point $(x,y)$ is defined over a feature space $X$ and a label space $Y=\{0, \dots, N-1\}$, while there are $K$ clients, each with access to local dataset $D_i=\{(x_i,y_i)\}$. (The above notations will be used for the rest of the paper unless otherwise stated.)

In the case of label distribution skew, each client only has access to a subset of all data classes. To better control and quantify data heterogeneity, we borrow the heterogeneity index from \cite{curse}:
\begin{equation}\label{hi}
    HI(n_c) = 1-\frac{n_c-1}{N-1},
\end{equation}
where $n_c$ represents the number of classes per client.

When training starts, the server distributes current global model parameters $w_G^t$ to $K$ clients in round $t$. Then, each client $i \in [K]$ runs an optimization algorithm such as stochastic gradient descent (SGD) for $E$ epochs with local dataset $D_i$ to obtain the local model parameter $w_i^t$, and sends its local update ${\Delta w}_i^{t+1} = w_i^t-w_G^t$ back to the server.
Finally, the server aggregates all updates to get the global model parameter $w_G^{t+1}$ for the next round:
\begin{equation}\label{eq1}
    w_G^{t+1} = w_G^t + \sum_{i=1}^K \alpha_i\cdot{\Delta w}_i^t,
\end{equation}
where $\alpha_i = \frac{|D_i|}{|D|}$ and $\sum_{i=1}^K\alpha_i=1$ in FedAvg algorithm \cite{FL}.
This iteration will continue until the global model achieves the desired performance.

\subsubsection{Backdoor Attack in Machine Learning}
\label{backdoorbackground}
Take the classification task as an example, the backdoor attack aims to inject a malicious pattern into the learning model which causes the model to misclassify the input stamped with a backdoor trigger into the target label. Formally, let $tg$, $(x,y)$, $(x^*,y_t)$ denote the trigger pattern, the original sample, and the trigger-pasted sample with the target label. The attacker's goal is to obtain the model $f$ with parameter $w_B$ which achieves both high attack success rate (ASR) and main task accuracy (MTA) by minimizing the loss function $L$ (e.g., cross-entropy loss) as follows:
\begin{equation}\label{eq2}
    \mathop{\min}_{w_B}(L(f(x),y) + L(f(x^*),y_t)).
\end{equation}
Then, during the inference phase, the attacker can attach the trigger to arbitrary input samples, causing it to be misclassified to the desired target class.

\begin{figure}[ht]
\centering
\includegraphics[scale=0.4]{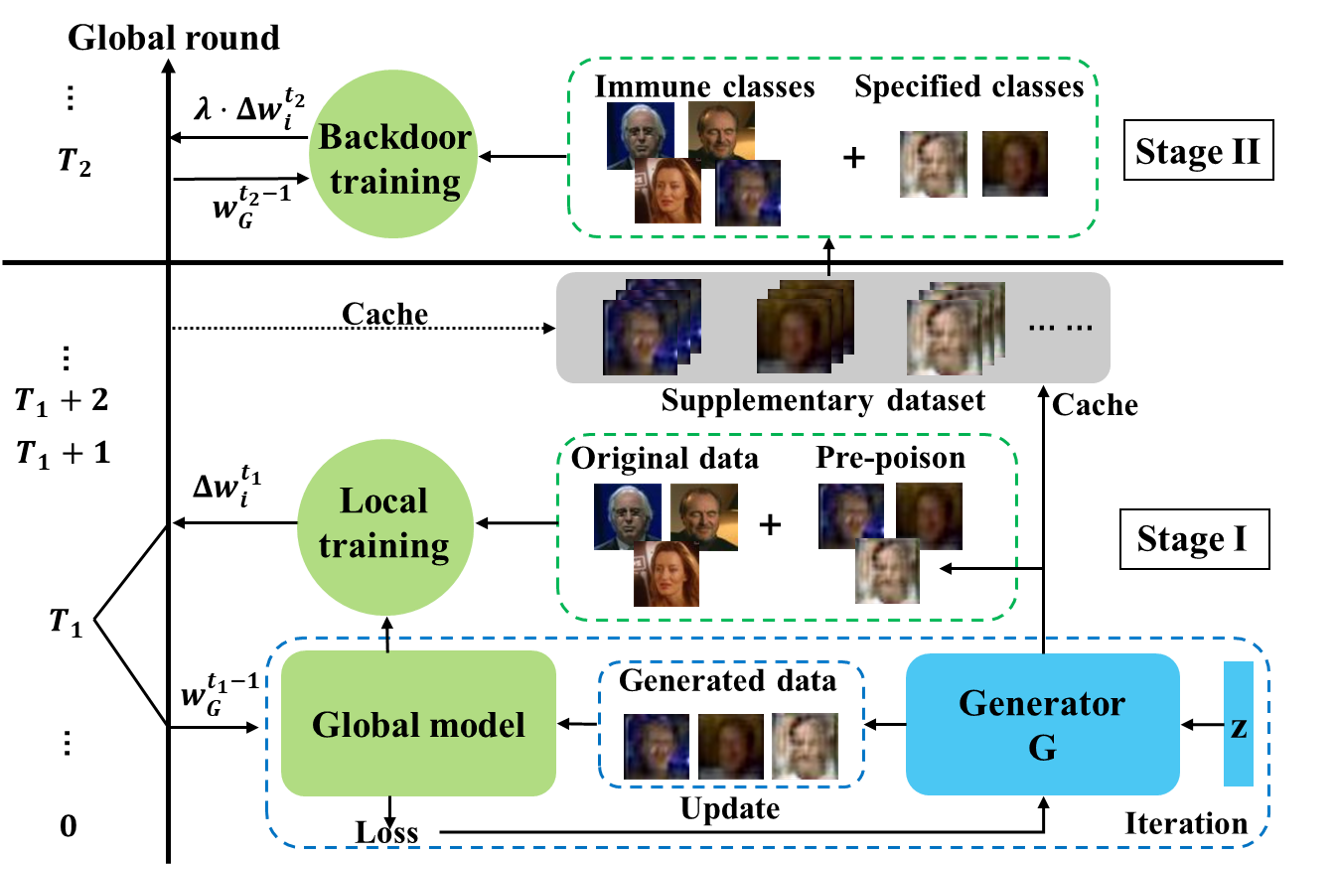}
\caption{The framework of our scheme. In stage I, the attacker obtains a supplementary dataset through the designed diverse data inference mechanism. In stage II, the attacker performs backdoor training against specified source classes within several rounds before the global model converges.}
\label{fig3}
\end{figure}

\subsection{Threat Model}
\subsubsection{Assumptions}
The main security threats faced by FL can be divided into two categories: model performance attacks and data privacy attacks, which may originate from both the server and the client. Some existing works assume a server-side attacker, who can steal private data through collected updates or directly manipulate the aggregated global model, while we consider a more challenging scenario: the attacker is on the client and the data distribution among all clients is label distribution non-IID, i.e., each client only has a subset of all classes as described in section \ref{pre1}.

\subsubsection{Adversarial Objectives}\label{metric}
As shown in Figure \ref{fig3}, there are two main and sequential goals throughout the attack process: private data inference and source-specified backdoor attacks. In the first stage, the attacker aims to generate high-quality samples of diverse data classes from the global model (especially those not locally available) to improve its local data distribution.
In the second stage, the attacker's goal is to inject an advanced backdoor that only works on the specified classes into the model, so as to determine the susceptibility and immunity of different clients to the backdoor trigger.

\subsubsection{Adversarial Capabilities}
Since the attacker is placed on the client side, its knowledge only includes the local data of partial classes and the global model parameters received in each round. 
The attacker's capability is also limited to planning malicious training locally through the above information and influencing the global model by uploading poisoned updates.

\begin{algorithm}[t]
\small
  \caption{GANs-based Diverse Data Inference}
  \label{alg1}
\begin{algorithmic}[1]
    \STATE {\bfseries Input:} the first-stage attack rounds $S_1$, local dataset $D$, number of GANs iterations $N$, number of local epochs $E$, local learning rate $\eta$, the target label $y_t$, the generator $\mathcal{G}(z,\theta)$, generator learning rate $\eta_g$.
    \STATE {\bfseries Output:} the supplementary dataset $D'$ of class $y_s$.
  
    \STATE $D' = \varnothing$
  
    \FOR{round $t$ in $S_1$}
        \STATE Receive the global model $w^t$ from the server
        \STATE $w_i^t \leftarrow w^t$
        \STATE 
        \STATE // Generation process
        \FOR{iteration $ n = 0,1,\dots,N-1$} 
            \STATE $ \boldsymbol{x_g} \leftarrow \mathcal{G}(z,\theta)$
            
            \IF{$L_G(w^t;\boldsymbol{x_g}) < threshold$}
                \STATE $D' \leftarrow D' \cup {\{\boldsymbol{x_g}\}}$
            \ENDIF
            
            \STATE $\theta \leftarrow \theta - \eta_g \nabla L_G(w^t;\{\boldsymbol{x_g}\})$
        \ENDFOR
        \STATE
        \STATE // Pre-poisoning process
        \FOR{local epoch $e = 0,1,\dots,E-1$}
        
            \FOR {each batch $\boldsymbol{b_i}=\{\boldsymbol{x},y\}$ of $D$}
                \STATE $\boldsymbol{b_i} \leftarrow \boldsymbol{b_i} \cup \{\boldsymbol{x_g},y_t\}$
                 \STATE $w_i^t \leftarrow w_i^t - \eta \nabla L(w_i^t;\boldsymbol{b_i}) $
            \ENDFOR
        \ENDFOR
        
        \STATE $\Delta w_i^t \leftarrow w^t - w_i^t$
        \STATE Send $\Delta w_i^t$ to the server
    \ENDFOR
    \RETURN $D'$
\end{algorithmic}
\end{algorithm}

\subsection{GANs-based Diverse Data Inference}\label{diversegen}
To solve the problem of missing data, we design a GANs-based diverse data inference mechanism to generate a supplementary dataset to improve the attacker's local data distribution. 
As we know, GANs consist of a generator $G$ and a discriminator $D$. $G$ is trained for generating desired samples by learning a mapping between distributions, e.g., from random Gaussian noise $z$, to the sample $x$ in real-word distribution $p_{real}$. While $D$ is trained to distinguish between real images and those generated by $G$. For vanilla GANs, the overall objective function can be formulated as follows:
\begin{equation}
\begin{split}
     \mathop{min}_G\mathop{max}_D V(G,D) & = \mathbb{E}_{x\sim p_{real}(x)}[\log D(x)] \\
    & + \mathbb{E}_{z\sim p_z(z)}[\log(1-D(G(z)))].
\end{split}
\end{equation}

However, due to the lack of real data, the attacker cannot locally train a discriminator to give effective feedback to the generator. To address this issue, a key insight is to regard the global model as a discriminator, since classification can be seen as a more specific type of discrimination \cite{cls_dis}. 
Different from the existing data inference attacks,
we introduce the data-free solution into FL and craft the generator loss function from two aspects of authenticity and diversity.

\textbf{Authenticity-oriented loss functions}. 
Let $\boldsymbol{y}^{sm} = D(\boldsymbol{x}) \in \mathbb{R}^N$ be the softmax output of the discriminator (i.e., the current global model) for input sample $\boldsymbol{x}$ and $\boldsymbol{y}^{oh} \in \mathbb{R}^N$ be the one-hot vector where $\boldsymbol{y}^{oh}_i = 1$ if $i = argmax_j\{\boldsymbol{y}^{sm}_j\}$. Then we can define the one-hot loss \cite{dfkd1}:
\begin{equation}
    L_{oh} = H_{ce}(\boldsymbol{y}^{sm},\boldsymbol{y}^{oh}) 
            = -\sum_{i=0}^{N-1} \boldsymbol{y}^{oh}_i \log\boldsymbol{y}^{sm}_i,
\end{equation}
where $H_{ce}$ represents the cross-entropy loss function. The one-hot loss function can encourage the softmax outputs of generated images by the global model to be close to the one-hot vectors, in other words, the generated images can be classified into any class with high confidence.

Besides, we also exploit the intermediate feature $\boldsymbol{f_i}$ of the input generated sample $x_i$ extracted by the global model based on the insight that $x_i$ with more significant $\boldsymbol{f_i}$ values is more likely to belong to the distribution of the original training dataset. Therefore, we define a feature significance loss function as follows:
\begin{equation}
    L_{fs} = - \log(\frac{\Vert \boldsymbol{f_i}\Vert_2}{dim(\boldsymbol{f_i})}),
\end{equation}
where $dim(\boldsymbol{f_i})$ is the dimension of vector $\boldsymbol{f_i}$.

\textbf{Diversity-oriented loss functions}. 
Here, we design the loss functions from the perspective of class diversity and sample diversity respectively.

Firstly, we introduce the information entropy to balance the generation probability of each class. In particular, let $\overline{\boldsymbol{p}}$ be the element-wise average vector of $\boldsymbol{y}^{sm}$ (as defined above) for all input samples, we define the information entropy as:
\begin{equation}
    L_{ie} = -H_{ie}(\overline{\boldsymbol{p}}) 
            = -\sum_{i=0}^{N-1} \overline{p_i} \log\overline{p_i},
\end{equation}
where $H_{ie}$ represents the information entropy function. By maximizing the information entropy of $\overline{\boldsymbol{p}}$, we force the generator to generate samples across all classes.

Besides, we note that mode collapse is a common problem in GANs as well as in existing data inference attacks \cite{deepmodel, poisongan} against FL. Therefore, inspired by \cite{modeseeking}, we design a mode-seeking function to promote the sample diversity within each class:
\begin{equation}
    L_{ms} = max_G (\frac{\Vert D(G(\boldsymbol{z_1}))-D(G(\boldsymbol{z_2}))\Vert_2}
    {\Vert \boldsymbol{z_1}-\boldsymbol{z_2} \Vert_2}),
\end{equation}
where $\boldsymbol{z_1}$ and $\boldsymbol{z_2}$ represent two different latent vectors. Note that 
we replace the original distance between $G(\boldsymbol{z_1})$ and $G(\boldsymbol{z_2})$ in the numerator because the output of the global model (serving as $D$) can indicate the similarity between the generated samples better than the original vector difference of them.

\textbf{Overall generator loss functions}. 
Finally, we obtain the overall loss function by combining the aforementioned items:
\begin{equation}
    L_G = L_{oh} + L_{fs} + \alpha (L_{ie} + L_{ms}),
\end{equation}
where $\alpha$ is a hyper-parameter balancing the two aspects of loss, which can be adjusted for different datasets.

\begin{figure}[ht]
\centering
\includegraphics[width = 3.6in]{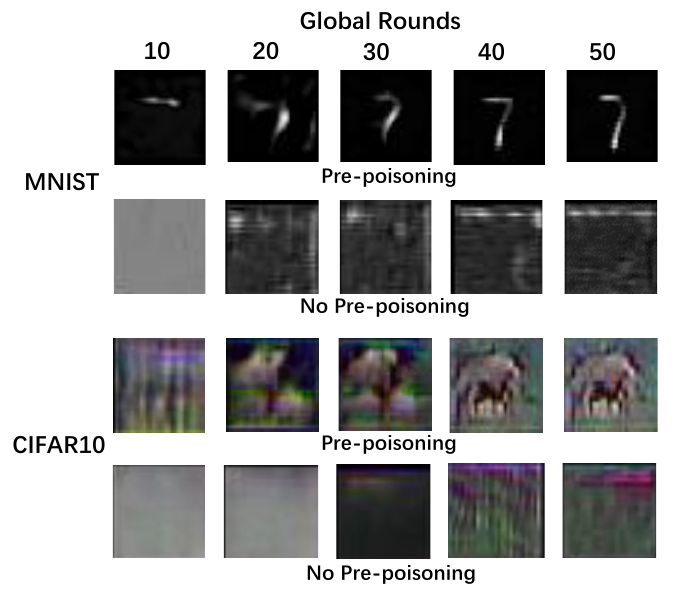}
\caption{Comparison of the effect of GANs-based privacy inference with and without pre-poisoning.}
\label{prepoison}
\end{figure}

\textbf{Pre-poisoning strategy}.
By designing the loss function, we can greatly enhance the performance of the generator, but this is far from enough in the FL settings. 
It should be noted that for the global model, the feature spaces of the negative samples in the classification and the discrimination task are completely different.
Specifically, we use $y_i^{real}$ and $y_i^{fake}$ to represent the samples of class $y_i$ from the real distribution and the generator $G$. For $y_i^{real}$, the negative samples should be $y_i^{fake}$ in the discrimination task while coming from the other classes $ Y \setminus \{y_i\}$ in the FL classification task. Therefore, if we simply use the global model in each round as the discriminator, the generator cannot learn more information about $y_i$, resulting in poor generation results as shown in Figure \ref{prepoison}. 
To solve this problem, we leverage a pre-poisoning strategy to actively inject a small number of generated samples that are mislabeled as the backdoor target class, as the negative samples in each training round. With this pre-poisoning strategy, the quality of generated data will be significantly improved and then the stealthiness and robustness of proposed backdoor methods is also improved.


After $G$ is updated in each round, the generated samples that reach the desired generator loss threshold will be cached to form a supplementary dataset in preparation for the next round of poisoning and the backdoor training in the second stage.

\subsection{Source-Specified Backdoor Learning}
Through continuous poisoning and generation in the first stage, the malicious client obtains a supplementary dataset which helps relieve the heterogeneity constraints and reduce the gap between local and global data distribution. After that, attackers can launch more effective and sophisticated backdoor attacks. As an example, we propose a source-specified backdoor learning (SSBL) strategy considering a realistic vicious competition situation in FL as mentioned in \ref{intro}.

Specifically, the attacker first specifies the victim class(es) out of all $N$ classes (perhaps from his competitors). Let $D_s = \{(x_s, y_s)\}$ and $D_{ns} = \{(x_{ns}, y_{ns})\}$ denote the datasets of the specified and non-specified classes respectively. After that, the backdoor training is carried out following three metrics:

\begin{itemize}
    \item \textbf{The attack success rate (ASR)} denotes the percentage of samples from $D_s$ with trigger patterns that are successfully classified as the target label (denoted as $y_t$) by the infected global model.
 
    \item \textbf{The main task accuracy (MTA)} denotes the accuracy of clean samples from all classes predicted by the infected global model.
    
    \item \textbf{The backdoor score (BS)} indicates the discrimination of the attack, i.e., the trigger only works on the specified classes, but has no effect on the non-specified ones:
    \begin{equation}\label{bs}
    \begin{split}
        BS =&  2\cdot S_1S_2/(S_1+S_2) \\
        s.t. \quad &S_1 = \mathbb{E}_{(x,y)\in D_s} \mathbb{I}\{f(x^*) \neq y\} \\
        &S_2 =  \mathbb{E}_{(x,y)\in D_{ns}} \mathbb{I}\{f(x^*) = y\}, \\
    \end{split}
    \end{equation}
     where $\mathbb{I}(\cdot)$ is the indicator function. $\mathbb{I}(A) = 1$ if and only if event `A' is true. 
\end{itemize}

To this end, we propose an objective function composed of three components for the backdoor training:
\begin{itemize}
    \item {$L(f(x^*_s),y_t)$}: for the specified class samples $x_s$  pasted with the trigger, we use the cross entropy of the prediction of $f$ with the backdoor target label $y_t$.
    \item {$L(f(x^*_{ns}),y_{ns})$}: for the non-specified class samples $x_{ns}$ pasted with the trigger, we use the cross entropy of the prediction of $f$ with their original label $y_{ns}$.
    \item {$L(f(x),y)$}: the original loss function for all clean data $(x,y)$.
\end{itemize}

Combining the above items, we define the source-specified backdoor loss function $L_{cs}$ as:
\begin{equation} \label{backdoorloss}
    \begin{split}
        L_{cs} = L(f(x^*_s),y_t) &+ \beta_1 \cdot L(f(x^*_{ns}),y_{ns})\\
        & + \beta_2 \cdot L(f(x),y),
    \end{split}
\end{equation}
where $\beta_1$ and $\beta_2$ are the hyper-parameters to trade-off between the loss items, and we empirically set $\beta_1 = \beta_2 = 0.1$ in this work.

Finally, to enhance the impact of malicious updates uploaded, we explicitly amplify the malicious update $\Delta w_B$ by a factor of $\lambda$ before uploading, the effect of which will be discussed in the experiment section.


\section{Experiments} \label{experiment}
\subsection{Experimental Setup}
Our work focuses on image classification tasks on three widely used datasets: MNIST, CIFAR10, and YouTube Aligned Face (YAF for short). We adopt Alexnet \cite{alexnet} as the initial global model and conduct FL over $K=100$ clients for 500 rounds. Before training begins, we perform data augmentation (such as rotation, cropping, and flipping) on the datasets to ensure that each client has at least 1000 samples. In each global round, only 10 clients are selected and trained for 2 local epochs. Models are optimized using SGD with a batch size of 32, and we use an initial learning rate of 0.01 with a decay schedule parameter of 0.95 every 20 global epochs.

\textbf{Label distribution non-IID data partitions.}
To simulate the situation of label distribution skew, we conduct label partitions following several previous works \cite{noniid1,noniid2}. 
For all $K$ clients, each with $n_c$ classes of data, we divide all training data into $K \cdot n_c$ shards, and each client will be randomly assigned $n_c$ shards from different classes.
In the experimental evaluation, we use the heterogeneity index defined in Equation \ref{hi} to control the degree of label distribution skew and all clients share the same heterogeneity index.

\textbf{Attack settings.}
We assume that the attacker controls 10\% of the clients, and at least one malicious client will be chosen in each global round. 
For the first-stage data inference, the generator will be optimized with SGD for 1,000 iterations with a batch size of 128. We use an initial learning rate of 0.01 with a decay schedule parameter of 0.9 every 50 iterations. 
In the backdoor training phase, $n_s$ classes will be specified as the victim, which, WLOG, will be misclassified into the first class of the dataset when the trigger is attached. Triggers are random pixel blocks and the size is $4 \times 4$ for MNIST and $8 \times 8$ for the other two datasets. After local backdoor training, we follow the strategy in \cite{ba1,ba2,poisongan}, allowing malicious updates to be amplified by $\lambda$ times before uploading to the server.
By default, we set the heterogeneity index as 0.6, the number of specified classes $n_s=3$, and the amplification factor $\lambda=3$.

\subsection{Effectiveness of GANs-based Diverse Data Inference}
\label{exp1}
According to \cite{curse}, date heterogeneity reduces the backdoor effectiveness and also challenges the design of a good attack strategy. 
In this part, we first present the results for data inference, and then further evaluate the improvements our method brings to regular backdoor attacks \cite{ba1,poisongan} under different heterogeneity indices.

\begin{figure}[ht]
\centering
\includegraphics[width = 3.4in]{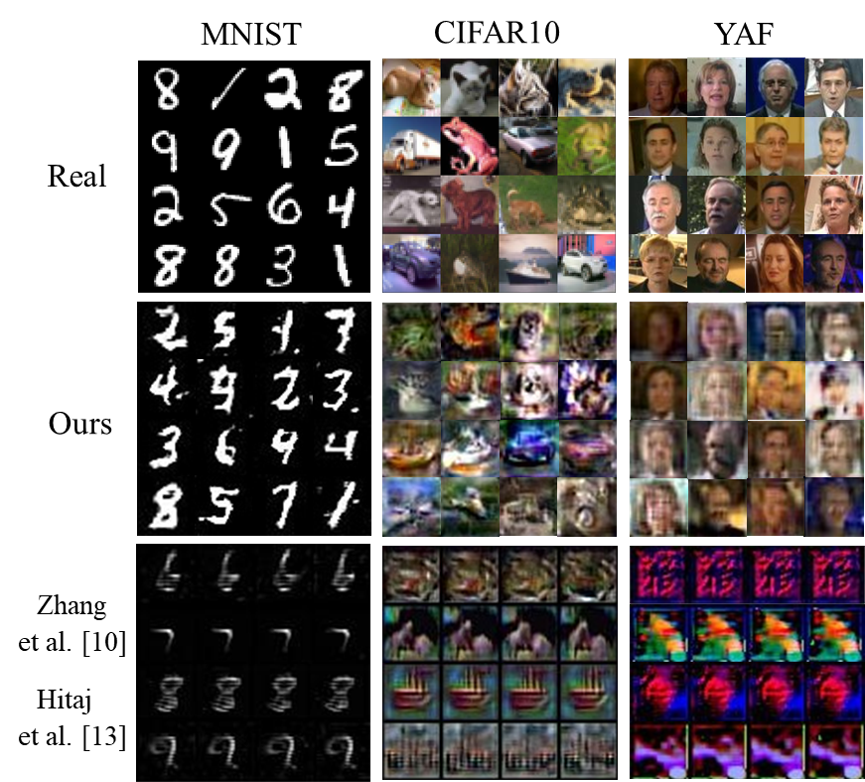}
\caption{Generated results comparison on MNIST, CIFAR10, and YouTube Aligned Face datasets. Note that methods in \cite{deepmodel} or \cite{poisongan} can only generate similar samples from a single class, while we display their different results only for comparison.
}
\label{reverse_compare}
\end{figure}

\textbf{Visualization.} In Figure \ref{reverse_compare}, we visualize the results of data inference and compare them with existing methods. As we expected, our approach performs better in generating samples with both authenticity and diversity, while the diversity is not only reflected in generating different classes each time but also in the difference of samples in each class. In contrast, the method in \cite{deepmodel} and \cite{poisongan} can only recreate a single class where all samples are similar. 
We also briefly demonstrate the effectiveness of the pre-poisoning strategy in Figure \ref{prepoison} and found that it can effectively promote faster and more stable generator training. 
An experimental finding is that the interpolation-based generator has better visual effects than the deconvolution-based one (e.g., DCGAN \cite{dcgan}), because the latter tends to show the checkerboard artifacts \cite{checkboard}.

\begin{figure*}[ht]
    \centering
    \includegraphics[scale=0.62]{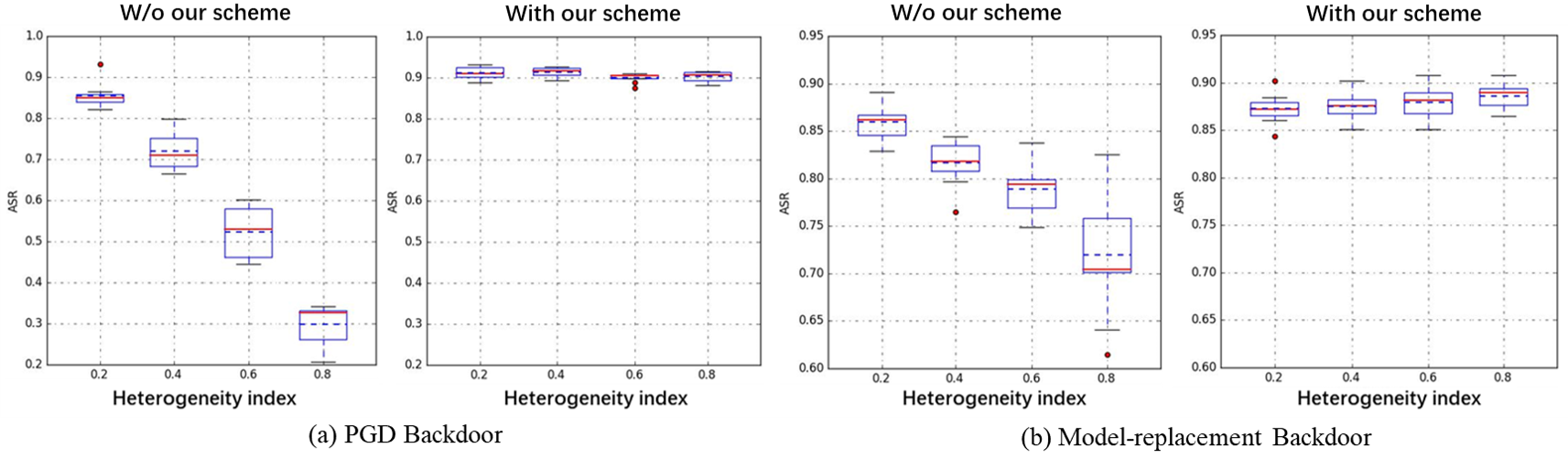}
\caption{Performance comparison of existing backdoor attacks \cite{ba1, tailattack} with and without our proposed data inference scheme under different heterogeneity indices.}
\label{normalbackdoor}
\end{figure*}

\textbf{Improvement.} In order to more intuitively understand how data heterogeneity affects backdoor attack effectiveness and to what extent our data inference methods can help attackers alleviate the limitations, we conducted a set of experiments by varying the heterogeneity index. 
We choose the typical model-replacement backdoor attack in \cite{ba1} and PGD backdoor attack in \cite{tailattack} as the baselines for CIFAR10 and MNIST, respectively. The data poisoning rate for malicious clients is set to 50\%.
We run 10 times experiments for each heterogeneity index, picking a different backdoor target class each time, and use box-and-whisker plots to report the ASR in Figure \ref{normalbackdoor}. We can easily find the larger the heterogeneity index, the more ASR drops. However, when combined with our data inference scheme, the ASR becomes stable and can be increased by 20\%--60\% in the case of extreme heterogeneity.
We believe that there are two main reasons. The first is that in the absence of data classes, the trigger feature can not generalize well and only work on the data classes that have participated in the backdoor training. Another reason is that benign clients also face the heterogeneity problem and tend to produce overfitting updates, which will further suppress the backdoor feature injected by the attacker.

To sum up, it is effective and necessary to introduce our proposed diverse data inference method for backdoor attackers in non-IID FL scenarios.

\subsection{Effectiveness of SSBL}
\begin{figure*}[htbp]
    \centering
    \includegraphics[scale=0.64]{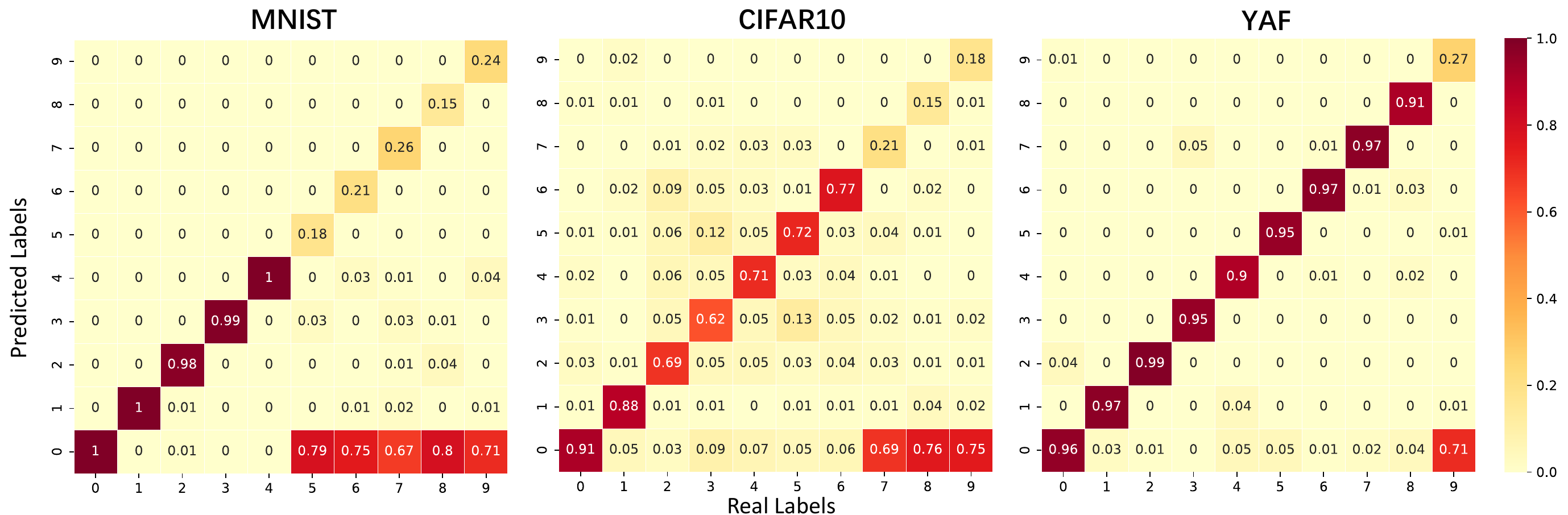}
\caption{Attack effect on MNIST, CIFAR10, and YAF datasets when the numbers of specified classes are 5, 3, and 1, respectively. We use a series of heatmaps to show the confusion matrices of the model injected by our backdoor when classifying the trigger-pasted images. As we can see, only the specified class are misclassified as the target.}
\label{heatmap}
\end{figure*}

The ``source-specified'' property of our attack not only requires a high probability of misclassification for specified classes (ASR) but also needs to ensure that the accuracy of non-specified classes is not affected. 
In order to visualize this kind of ``differential treatment”, we use the confusion matrix in Figure \ref{heatmap} to show the effectiveness of SSBL. 
WLOG, We specified 5, 3, and 1 victim classes for MNIST, CIFAR10, and YAF datasets, respectively, and set the first class as the backdoor target. Note that we have no restrictions on the choice of the target because the data from the target class are not involved during the attack process. 
As shown in the heatmaps, our SSBL is able to greatly reduce the model performance on the specified class, while the influence on non-specified classes is much smaller (even with no effect on some classes).
To reflect the effect more concretely, Table \ref{table1} gives a detailed comparison of the performance of our SSBL on three datasets when specifying different numbers of victim classes.
Here, we use $\Delta$MTA to represent the value by which the backdoor attack reduces the main task accuracy. Existing work \cite{noniid1} has shown that FL accuracy will be significantly reduced in non-IID data scenarios, so it is more useful to examine the variation when studying the impact of our attacks on model performance.
From Table \ref{table1}, we can see as the number of specified classes increases, the MTA is almost unaffected, which is consistent with the general characteristics of backdoor attacks. However, both the ASR and BS show a downward trend, which seems counter-intuitive, because the regular backdoor can be regarded as a special case of our SSBL with all classes specified. We conjecture that the reason lies in the adversarial relationship between the specified and non-specified classes. Specifically, in a traditional backdoor attack or the SSBL with a single class specified, this adversarial relationship is overwhelming, but in SSBL with half of the specified class, this relationship becomes balanced, making it more difficult for a trigger to find the boundary where it can distinguish them.

\begin{figure*}[htbp]
    \centering
    \includegraphics[scale=0.55]{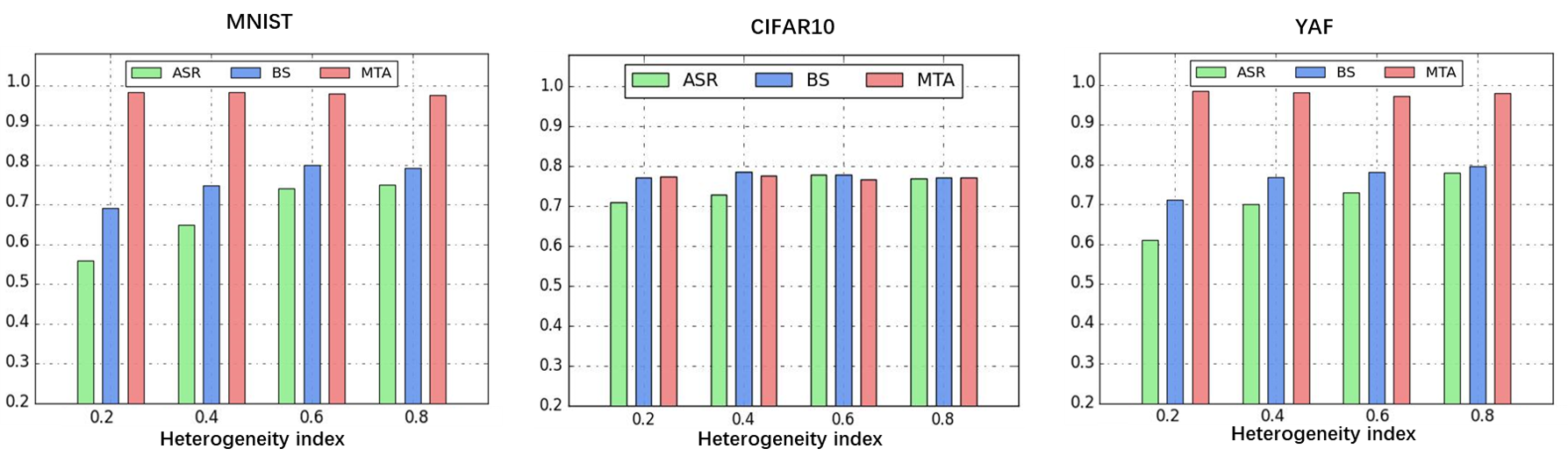}
\caption{Performance of SSBL on three datasets under different heterogeneity indices.}
\label{ablation_noniid}
\end{figure*}
\begin{figure*}[htbp]
    \centering
    \includegraphics[scale=0.55]{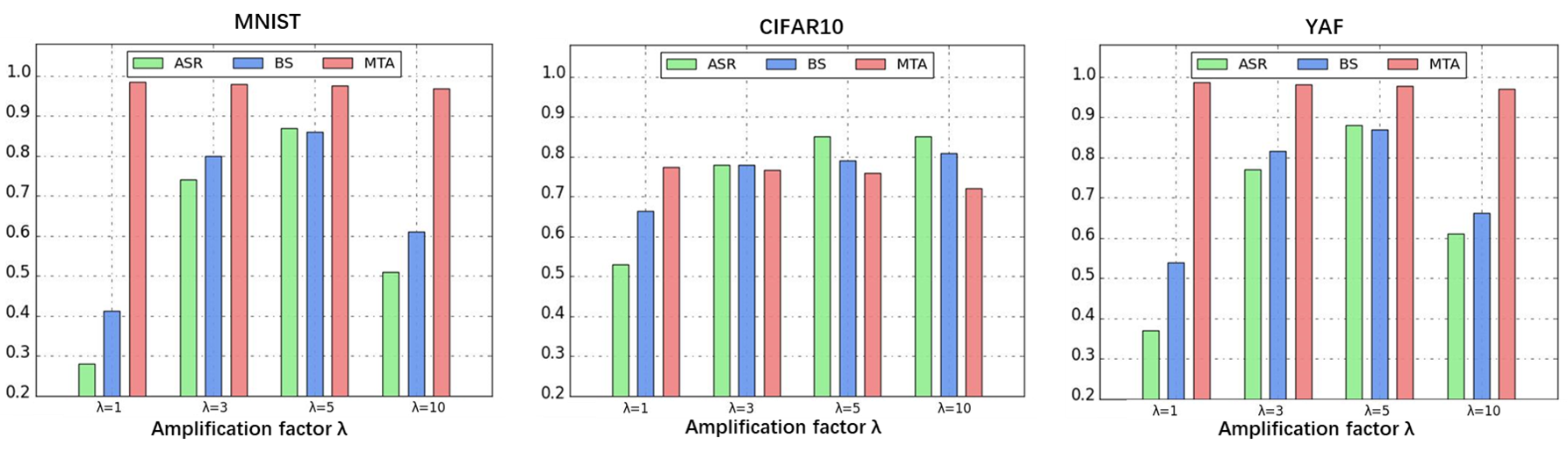}
\caption{Performance of SSBL on three datasets with different amplification factors.}
\label{ablation_lambda}
\end{figure*}

\begin{table*}
\renewcommand\arraystretch{1.5}
\begin{center}
\caption{the performance of the SSBL on three datasets when specifying different numbers of source classes}
\label{table1}
\setlength{\tabcolsep}{3mm}{
\begin{tabular}{ccccc|ccc|ccc}
\hline
\multicolumn{1}{l}{\multirow{2}{*}{Dataset}} & $n_s$ & \multicolumn{3}{c}{1} & \multicolumn{3}{c}{3} & \multicolumn{3}{c}{5}\\
\cline{2-11} 
\multicolumn{1}{l}{} & Metrics   & $ASR$           & $\Delta MTA$  & $BS$            & $ASR$  & $\Delta MTA$  & $BS$            & $ASR$  & $\Delta MTA$  & $BS$                       \\ \hline
\multicolumn{2}{c}{MNIST}                                & \textbf{0.81} & 0.01 & \textbf{0.86} & 0.76 & 0.03 & 0.79          & 0.74 & 0.03 & \multicolumn{1}{c}{0.77} \\
\multicolumn{2}{c}{CIFAR10}                              & \textbf{0.78} & 0.03 & 0.81          & 0.75 & 0.04 & \textbf{0.82} & 0.65 & 0.04 & 0.69                     \\
\multicolumn{2}{c}{YAF}                                  & \textbf{0.71} & 0.05 & \textbf{0.81} & 0.68 & 0.05 & 0.78          & 0.59 & 0.05 & 0.74                     \\ \hline
\end{tabular}}
\end{center}
\end{table*}

\subsection{Ablation Study}
\subsubsection{Heterogeneity Index}
We have analyzed the impact of heterogeneity index on traditional backdoor attacks in section \ref{exp1}. In this section, we comprehensively examine the performance of SSBL through three metrics ASR, MTA, and BS (as defined in section \ref{metric}) by setting different heterogeneity indices.

There are two points that we can learn from Figure \ref{ablation_noniid}. First, the ASR increases with the higher degree of heterogeneity, which shows that our method can even take the advantage of data heterogeneity and has strong adaptability to non-IID data FL scenarios. Second, there exists an adversarial relationship between the specified and non-specified classes (which is also reflected in the experiment in Table \ref{table1}). For example, in the results of CIFAR10, there is no significant change in BS as ASR increases, which, by definition, we can infer is due to a decline in the accuracy of non-specified classes.

\subsubsection{Amplification Factor $\lambda$}
Following most of the existing works \cite{ba1,ba2,poisongan}, we introduce a parameter $\lambda$ to control the amplification of the backdoor update. In a more general scenario, the selection of $\lambda$ mainly needs to consider the total number of clients, the proportion of attackers, and the frequency of attackers being selected.
Here, we chose the four cases where $\lambda$=1, 3, 5, and 10. The results are shown in Figure \ref{ablation_lambda}.
It is worth noting that since we can achieve a more effective backdoor attack with the help of our data inference mechanisms, the required amplification factor is much smaller than existing works (even up to 40 in \cite{poisongan}).
As we can see, update amplification is very important for the attack effectiveness. For example, a simple 3x amplification can increase the ASR by as much as 50\% on MNIST. However, excessive amplification may damage the performance of the global model, leaving the backdoor attacker exposed.
In addition, when $\lambda$=10, both MNIST and YAF datasets show a decrease in ASR, which may be due to the effect of the penalty term in our source-specified loss function being amplified at the same time.

\subsection{Stealth Analysis of SSBL Strategy Against Defense Method}
Although there have been many studies on defense strategies against backdoor attacks in FL, such as FoolsGold \cite{foolsgold}, FLGUARD \cite{flguard}, and FLAME \cite{flame}, few of them are applicable to label distribution non-IID scenarios. 
There are two main reasons.
First, most defense methods \cite{foolsgold, flguard, flame} are based on the idea of anomaly detection, i.e., to discover and eliminate malicious updates through similarity comparison or clustering, etc. However, there are naturally significant differences in client updates under non-IID scenarios\cite{survey_anomalydetection}, making such methods invalid.
Second, the secure aggregation mechanism \cite{sa} adopted by many FL systems makes it impossible for defenders to examine the clients' updates.
Therefore, we focus on post-training backdoor detection to evaluate the stealth of our attack in this work. 

\subsubsection{Stealth against Neural Cleanse}
\begin{figure}[tb]
\centering
\includegraphics[scale=0.48]{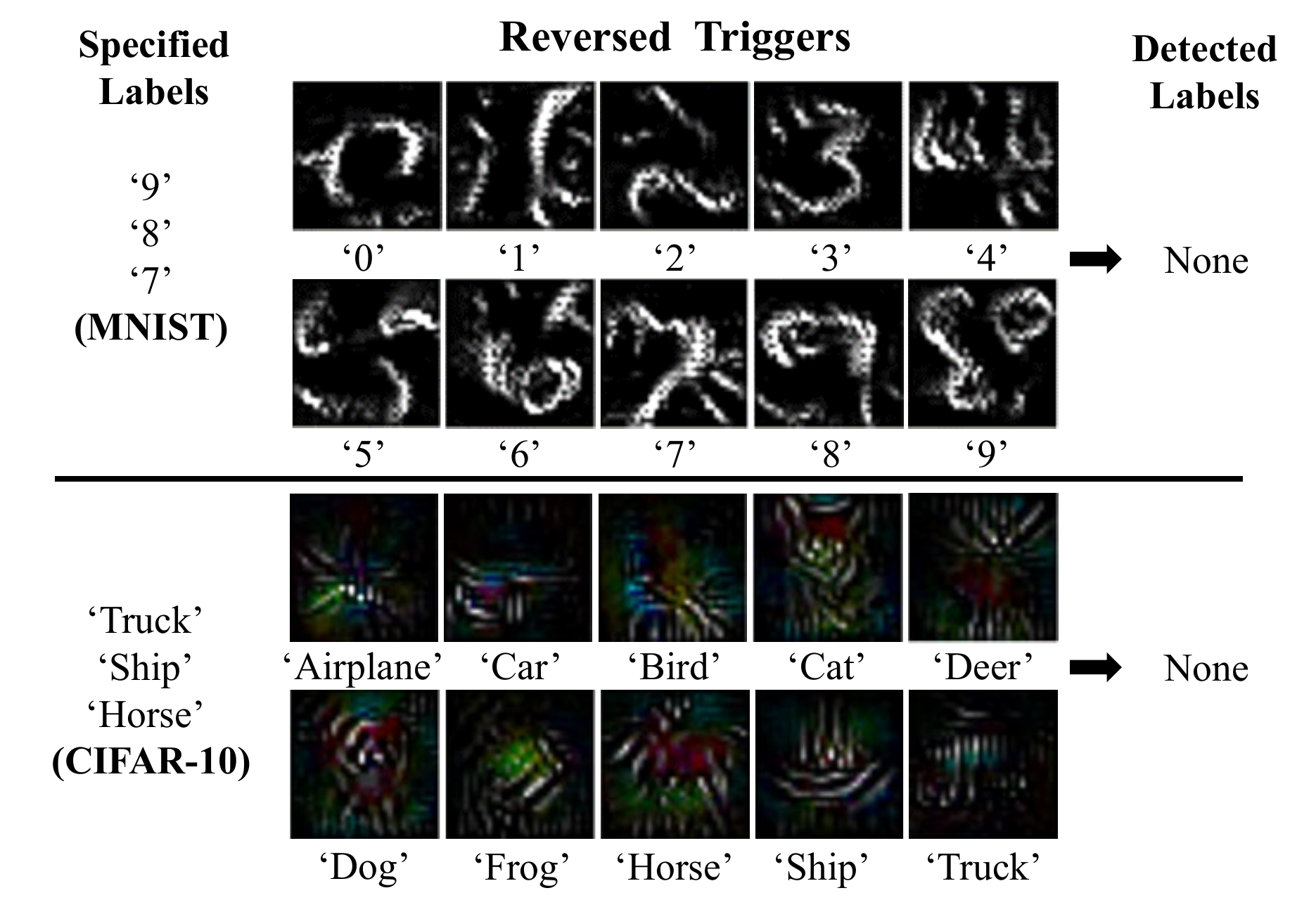}
\caption{The trigger reverse results by Neural Cleanse on MNIST and CIFAR10. The backdoor target class is set to class 0.}
\label{detection}
\end{figure}

We first use the popular Neural Cleanse (NC) method \cite{neuralcleanse} to detect the global model that has been attacked by our SSBL strategy in FL.
The NC leverages a gradient descent-based approach to generate a possible trigger for each class. After that, the NC uses the Median Absolute Deviation (MAD) as the outlier detection algorithm to detect the reversed triggers and identify the backdoor target class.
We use class 0 as the backdoor target in MNIST and CIFAR10 datasets, and specify 3 classes as the victims. The trigger restoration effect is shown in Figure \ref{detection}. As we can see, for our source-specified attack, Neural Cleanse failed to restore the trigger or distinguish the target class. In addition, Table \ref{table2} reports the detection results of NC under different $n_s$. It is intuitive that the fewer source classes are specified, the harder the attack is to detect.
\subsubsection{Stealth against Activation Clustering}
The Activation Clustering (AC) method \cite{activationclustering} uses the k-means algorithm to cluster the activations extracted from the last hidden layer for input images. By analyzing the silhouette score of the clustering results, the AC method can distinguish the clean and poisoned classes. 
As we can see from \ref{table2}, the AC method fails to detect most of the time but is more effective and more stable than the NC method. The reason may be that AC can identify anomalous samples based on clustering regardless of the number of specified source classes.
However, the AC approach has a serious drawback, i.e., the defender needs to access a large amount of training data including poisoning data, which is difficult in FL because of the violation of privacy protection.

\begin{table}[]
\caption{Average detection accuracy over 10 experiments of two defense methods when specifying different numbers of source classes. The last column represents results with no source class specified. All values are percentages.}
\label{table2}
\begin{center}
\renewcommand\arraystretch{1.5}
\begin{tabular}{cccccc}
\hline
Defense     & Dataset & $n_s$=1 & $n_s$=3 & $n_s$=5 & \textbackslash{} \\ \hline
\multirow{3}{*}{\begin{tabular}[c]{@{}c@{}}Neural \\ Cleanse\end{tabular}}  
& MNIST   & 0   & 20  & 40  & 100              \\
& CIFAR10 & 0   & 10  & 20  & 100              \\
& YAF     & 0   & 10  & 20  & 100              \\ \hline
\multirow{3}{*}{\begin{tabular}[c]{@{}c@{}}Activation\\ Clustering\end{tabular}} 
& MNIST   & 20  & 30  & 50  & 100              \\
& CIFAR10 & 10  & 30   & 40  & 100              \\
& YAF     & 20  & 20  & 40  & 100              \\ \hline
\end{tabular}
\end{center}
\end{table}

\subsubsection{Analysis of Possible Defense}
Our SSBL strategy narrows the attack scope on the backdoor source classes, making it more difficult to detect. 
But there are still some defenses that can be effective, especially when the defenders are aware of the source-specific nature of the backdoor attack. For example, the NC defender can reverse engineer the triggers of all possible source-target label pairs to detect the anomaly ones at the cost of a significant increase in computation \cite{neuralcleanse}.

Beyond that, more vigilant defenders can assume the model has been injected with a backdoor and go straight to backdoor elimination such as \cite{eliminate_backdoor,finepruning}. We speculate that this may be effective because our SSBL strategy does not consider optimizing for backdoor persistence. But it should be noted that such coarse-grained defense may come at the cost of degradation in the performance of the original model.

\subsection{Future Work}

\textbf{Defense for data inference.} 
Data inference utilizes the inherent characteristics of the DNN model, i.e., a well-trained model can memorize information of training data. In this sense, as long as the clients have white-box access to the global model, data inference attacks are unavoidable. 
Therefore, a defense idea is to restrict the client's access to the global model during the training phase and only provide an interface for uploading data, so as to prevent the client-side attacker from stealing additional information.

\textbf{Cross-device FL scenario}. In cross-device FL, the clients consist of many devices (sometimes millions), each with a small dataset. In this case, the attacker will face more extreme resource-constrained challenges. How to optimize our data inference mechanism and design new attack strategies (such as multiple client collusion) remains to be further explored in such a more complex distributed environment.

\textbf{Aggregation algorithms for non-IID FL.}
Algorithms such as FedProx \cite {fedprox} can speed up the convergence of the global model, which may facilitate the proposed data inference process. On the other hand, the modification of some algorithms to the local training process may also affect the backdoor effectiveness.

\section{Conclusion} \label{conlusion}
In this paper, we propose a privacy inference-empowered backdoor attack scheme for non-IID FL, which enables the client-side attacker to launch a more stealthy but effective backdoor attack under the label distribution skew scenario. 
Firstly, a GANs-based diverse data inference mechanism is proposed to generate private data, which uses a pre-poisoning strategy to improve the quality of generated data. 
Secondly, we further design a source-specified backdoor learning strategy, which allows the attacker to arbitrarily specify which classes are susceptible to the backdoor trigger. 
Extensive experiments are implemented based on MNIST, CIFAR10, and YouTube Aligned Face datasets to evaluate the effectiveness of the proposed methods. By our diverse data inference mechanism,
we can increase the ASR of regular backdoors by over 60\% when it is ineffective due to data absence. And the SSBL strategy can successfully evade state-of-the-art backdoor detection methods.
Overall, our scheme presents higher feasibility and flexibility and is compatible with other backdoor attacks to improve their effectiveness and stealthiness, which can provide strong support for future research on backdoor attacks in non-IID FL.

\section*{Acknowledgment}
This research work is funded by the National Nature Science Foundation of China under Grant No. 62202303, U21B2019 and U20B2048, Shanghai Sailing Program under Grant No. 21YF1421700, and CCF-AFSG RF20220018, and Defence Industrial Technology Development Program Grant No. JCKY2020604B004.

\normalem

\bibliography{ref.bib}
\bibliographystyle{IEEEtran}

\end{document}